\documentclass[journal]{IEEEtran}

\usepackage{cite}
\usepackage{amsmath,amssymb,amsfonts}
\usepackage{graphicx}
\usepackage{booktabs}
\usepackage{algorithm}
\usepackage{algorithmic}
\usepackage{textcomp}
\usepackage{xcolor}
\usepackage{url}
\begin{document}



\title{\texttt{MemoGuard}: An Adaptive Runtime for Guarding Against Memory Traps in Communication-Limited Robot Navigation}

\author{Rajat~Bhattacharjya\IEEEauthorrefmark{2}$^{\ast}$\thanks{$^{\ast}$ Equal contribution. R. Bhattacharjya is the corresponding author (e-mail: rajatb1@uci.edu).\\ Paper accepted at ESWEEK CODES 2026 in the Late Breaking Results (LBR)
track. Authors' version posted for personal use and not for redistribution. The definitive version of the paper will appear in IEEE Embedded Systems Letters.}, 
        Hyeonjong~Ju\IEEEauthorrefmark{3}$^{\ast}$\textsuperscript{\stepcounter{footnote}\thefootnote}, 
        Sing-Yao~Wu\IEEEauthorrefmark{2}, 
        Eli~Bozorgzadeh\IEEEauthorrefmark{2}, 
        Nikil~Dutt\IEEEauthorrefmark{2} \\
\IEEEauthorrefmark{2}Department of Computer Science, University of California, Irvine \\ \IEEEauthorrefmark{3} Department of Computer Science, Yonsei University, South Korea}


\maketitle
\begin{abstract}
Communication-limited robots in mission-critical scenarios such as disaster inspection and search-and-rescue must make
reliable onboard decisions without access to remote operators or high-capacity reasoning
services. Episodic memory reuse is an attractive low-cost fallback, but retrieval
similarity does not guarantee execution validity, i.e., a retrieved action may match the
current context yet be unsafe due to changed topology, insufficient battery margin,
or unreliable prior outcomes. We call such high-similarity but execution-invalid
episodes \emph{memory traps}. This creates a safety-efficiency design space where similarity
only reuse minimizes fallback cost but can be unsafe, while
always invoking local reasoning improves safety at high computational and energy cost. This paper presents \texttt{MemoGuard}, a
lightweight adaptive runtime that validates episodic memories
against topology, resource, and outcome contracts before reuse,
invoking fallback only when validation fails. In a graph-based
corridor-inspection simulator, \texttt{MemoGuard} reduces battery safety
violations by 76.6\% over similarity-only top-1 reuse while reducing fallback calls by 21.4\% over always reasoning. On an
NVIDIA Jetson AGX Xavier with local \texttt{llama3.2:3b} fallback
reasoning, this corresponds to 3.67 s and 36.97 J of avoided
fallback-reasoning overhead per trial. We open-source \texttt{MemoGuard} at \url{https://github.com/hetheiin/memoguard}.
\end{abstract}


\section{Introduction}

Mobile robots operating in mission-critical settings such as disaster inspection and search-and-rescue increasingly rely on operator-specified mission intent to guide autonomous behavior~\cite{bhattacharjya2025avery, cladera2025air}. In practice, however, these robots may experience degraded or intermittent communication with remote operators, cloud planners, or high-capacity reasoning services~\cite{bhattacharjya2025avery,nakanoya2023co}. During such intervals, the robot must make onboard decisions with limited compute, energy, and sensing reliability~\cite{bhattacharjya2025avery,7139494}. A natural fallback is case-based episodic reuse: retrieving a past execution episode that resembles the current situation and reusing or adapting the associated action sequence~\cite{aamodt1994case,rothfuss2018deep}.

\begin{figure}[t]
    \centering
    \includegraphics[width=\columnwidth]{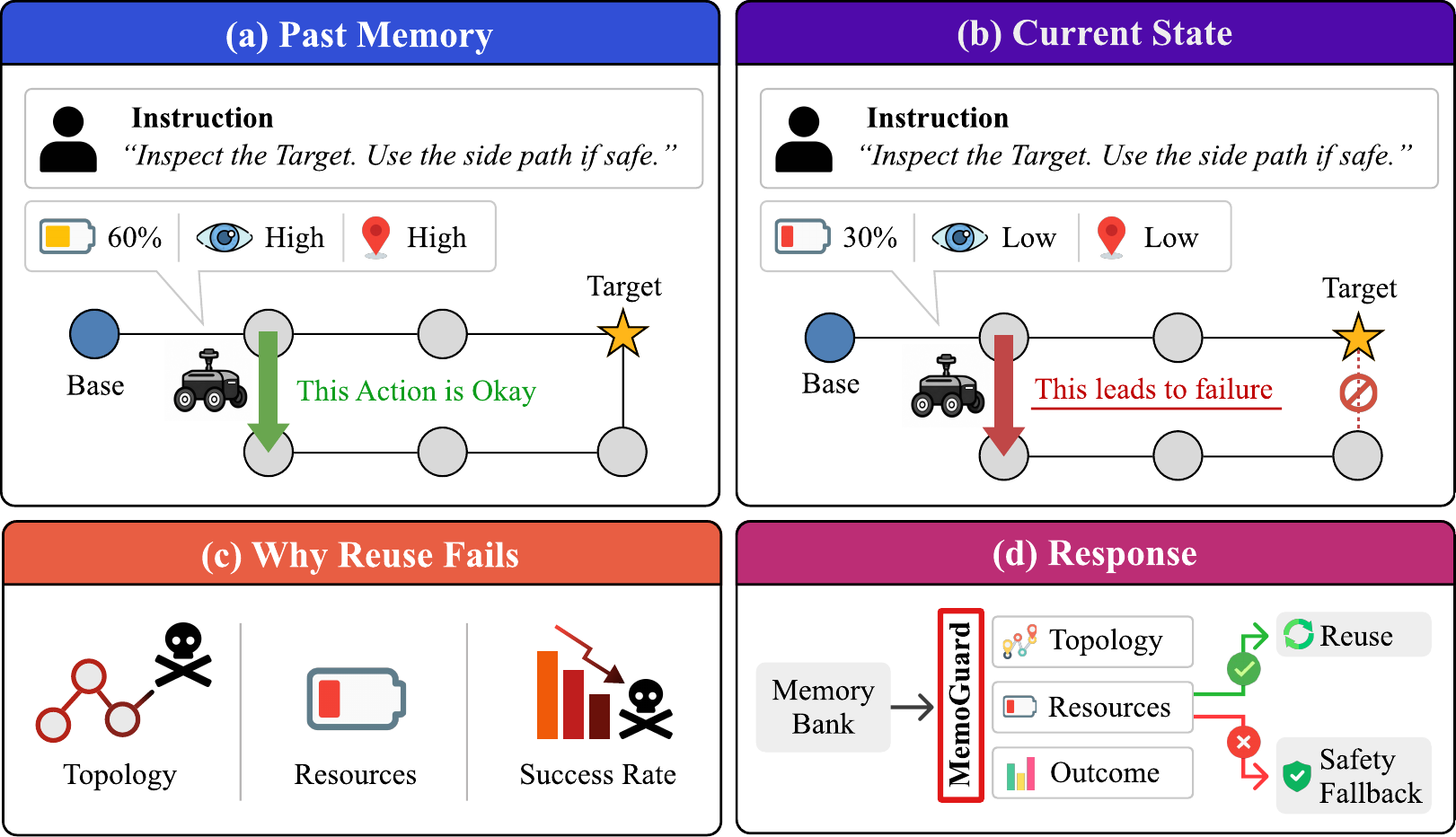}
    \vspace{-4ex}
\caption{\scriptsize Motivating memory trap under communication loss.
(a) A past memory records a successful action for the mission intent: inspect target \(T\) using the side path if safe.
(b) The current state has the same intent and similar local context, but changed topology, resource, and localization conditions make the remembered action unsafe.
(c) Similarity-based reuse fails by executing the stale action.
(d) \texttt{MemoGuard} rejects the unsafe memory by validating topology, resource feasibility, and outcome reliability before fallback.}
    \label{fig:motivating-memory-trap}
    \vspace{-4ex}
\end{figure}

This strategy is appealing for embedded robots because memory lookup and lightweight
validation can be much cheaper than invoking a local VLM~\cite{zheng2026freqcache} or
large reasoning module~\cite{check}. However, embodied memory reuse is not only a
retrieval problem. Fig.~\ref{fig:motivating-memory-trap} illustrates this problem in a
corridor-inspection mission: a robot may retrieve a past episode matching the current
intent, visibility, and blockage context, yet the remembered action may be invalid
because the side route is blocked, the battery margin is insufficient, or localization
confidence has degraded. A memory can be semantically relevant while its execution
assumptions no longer hold.


Prior case-based and episodic-memory approaches show that past experiences can guide
action selection~\cite{aamodt1994case,rothfuss2018deep}, but reuse in resource-constrained
settings is dominated by retrieval similarity while execution assumptions remain implicit.
This leaves a systems gap: before executing a retrieved action, the robot must verify
that those assumptions still hold under the current topology, resource trajectory, and
outcome history. We call high-similarity memories that fail this test \emph{memory traps}.


This paper presents \texttt{MemoGuard}, a lightweight runtime for contract-validated
episodic memory reuse in communication-limited embodied robots navigating the safety-efficiency design space. \texttt{MemoGuard}
treats each memory as an \emph{episodic action memory}: a compact record containing
the mission context, pre-state, remembered action, execution contract, and outcome
statistics. At runtime, \texttt{MemoGuard} validates candidates using hard feasibility
gates over topology, resource feasibility, and outcome reliability, invoking a
planner or local reasoning (LLM) fallback only when validation fails.



We make the following contributions:
\vspace{-1mm}
\begin{itemize}
    \item We identify \textbf{\emph{memory traps}} and design \texttt{MemoGuard}, a lightweight
    contract-validation runtime that guards against unsafe episodic reuse via hard
    feasibility gates and outcome-reliability scoring over a bounded supervisory
    action set.
    \item We evaluate \texttt{MemoGuard} in a graph-based corridor-inspection simulator
    across three topologies with blocked routes, altered viewpoint affordances, and resource
    constraints.
    \item We calibrate per-call fallback-reasoning overhead on an NVIDIA Jetson AGX Xavier, showing that reduced fallback frequency translates to 3.67~s and 36.97~J of avoided LLM fallback overhead per trial in \texttt{MODE\_30W\_ALL}.
\end{itemize}
\section{Memory Traps and Problem Formulation}
\label{sec:problem}
We now formalize the failure mode introduced in Fig.~\ref{fig:motivating-memory-trap}. We model communication-limited robot navigation as a supervisory decision problem in which a robot may either reuse a retrieved episodic memory or invoke a higher-cost planner/reasoning fallback. The core challenge is that retrieval similarity and execution validity are different properties: a memory may appear relevant while its remembered action is no longer executable.\\
\textbf{\textit{\underline{{(A) Episodic Action Memories:}}}} We model the operating environment as a topological graph \(G_t=(V_t,E_t)\), where
nodes represent locations (base, corridor, target, safe waypoint, alternate viewpoint)
and edges encode traversal cost, obstacle state, and local visibility. At decision
step \(t\), the robot state \(s_t\) includes the current node, target, battery,
localization confidence, communication state, mission intent, and local observations.
We focus on communication-limited intervals in which the robot cannot reliably access
remote operators, cloud planners, or high-capacity reasoning services.

The robot acts through a bounded supervisory action set:
\begin{equation}
\begin{aligned}
\mathcal{A} = \{&\texttt{follow\_planner},\ \texttt{inspect\_target}, \\
               &\texttt{return\_to\_safe\_waypoint}, \\
               &\texttt{inspect\_alternate\_viewpoint}, \\
               &\texttt{wait\_for\_recovery},\ \texttt{abort\_mission}\}.
\end{aligned}
\label{eq1}
\vspace{-1ex}
\end{equation}
These actions cover path following, target inspection, recovery, and mission
termination. Reasoning is not itself an executable action; the planner/reasoning
fallback selects one action from \(\mathcal{A}\).

An episodic action memory is a compact record of a prior execution event:
\begin{equation}
    m_i = (\iota_i,\, x_i,\, a_i,\, \kappa_i,\, \rho_i),
    \vspace{-1ex}
\end{equation}
where \(\iota_i\) is the intent context; \(x_i\) is the retrieval key (current node,
target, visibility, obstacle state, edge cost, battery range, localization confidence);
\(a_i\in\mathcal{A}\) is the remembered action; \(\kappa_i\) is the execution contract
(topological preconditions, viewpoint availability, expected action cost, path-risk
profile, and safety constraints); and \(\rho_i\) stores success and failure counts
for equivalent state-action memories.\\
\textbf{\textit{\underline{{(B) Top-$k$ Retrieval and Memory Traps:}}}} Given current state \(s_t\), the memory retriever first filters memories by intent context, current node, target, and mission phase, producing an eligible set \(\mathcal{M}_t\). It then ranks eligible memories by a retrieval similarity score \(S_{\mathrm{ret}}(s_t,m_i)\) and returns the top-\(k\) candidate set\footnote{Our evaluated prototype uses the lightweight \(k=1\) instantiation.}:
\begin{equation}
\mathcal{C}_t^{k}
=
\operatorname{TopK}_{m_i\in\mathcal{M}_t}
S_{\mathrm{ret}}(s_t,m_i).
\vspace{-1ex}
\end{equation}
A memory trap occurs when a retrieved memory appears similar to the current state but is invalid for execution. Let \(\Phi(s_t,m_i)\in\{0,1\}\) denote whether memory \(m_i\) is valid for reuse at state \(s_t\). We define the top-\(k\) memory-trap set as
\begin{equation}
\mathcal{T}_t^k =
\{m_i\in\mathcal{C}_t^k : \Phi(s_t,m_i)=0\}.
\end{equation}
Validity depends on whether the remembered action remains compatible with the current mission, topology, resource state, and outcome history:
\begin{equation}
\begin{aligned}
\Phi(s_t,m_i)=&~\Phi_{\mathrm{intent}}(s_t,m_i)
\wedge \Phi_{\mathrm{topo}}(s_t,m_i)\\
&\wedge \Phi_{\mathrm{res}}(s_t,m_i)
\wedge \Phi_{\mathrm{out}}(m_i).
\end{aligned}
\vspace{-1ex}
\end{equation}
Thus, a memory trap is not necessarily an irrelevant memory. It may match the current state semantically or locally while still being unsafe because its execution assumptions no longer hold, e.g., due to  changes in the physical environment or context.
\begin{figure*}[t]
    \centering
    \includegraphics[width=0.9\textwidth]{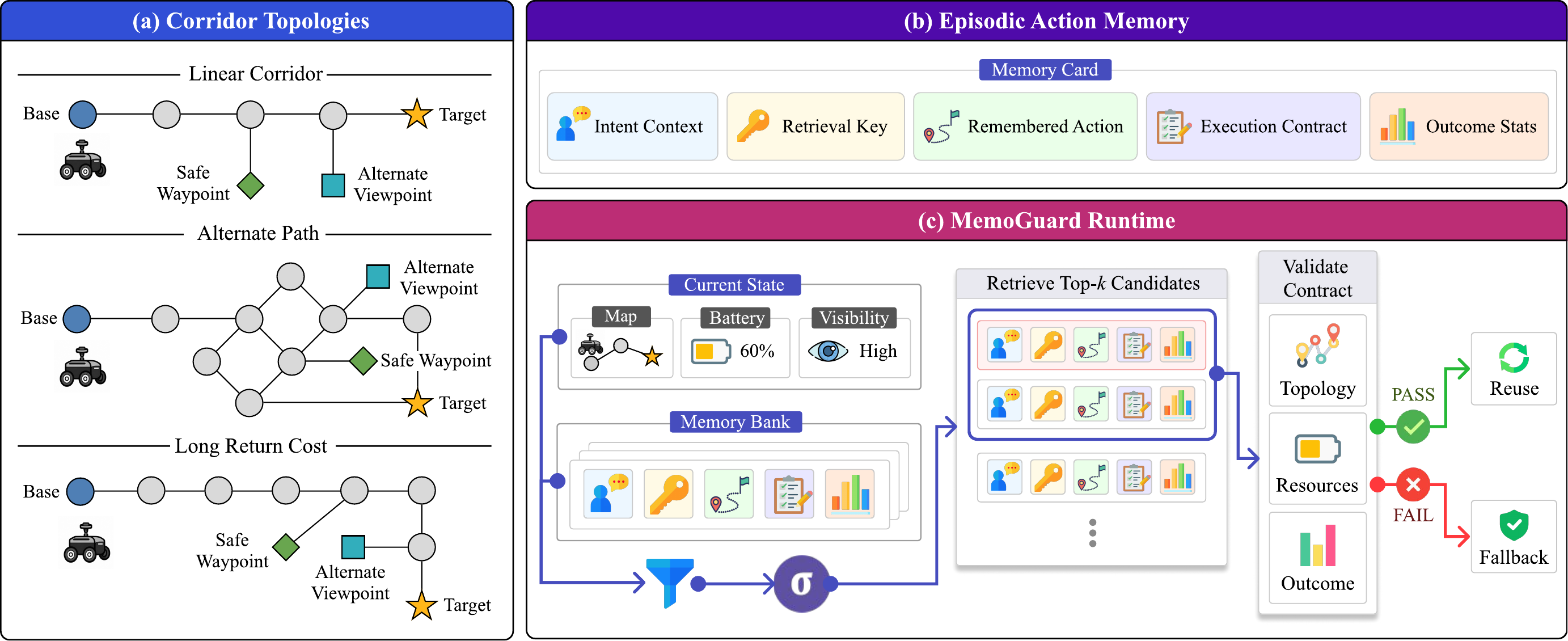}
    \vspace{-2ex}
    \caption{\scriptsize \texttt{MemoGuard} framework. 
    (a) Corridor topologies used in evaluation: Linear Corridor, Alternate Path, and Long Return Cost. 
    (b) Episodic action memory format, including intent context, retrieval key, remembered action, execution contract, and outcome statistics. 
    (c) Runtime flow: candidate memories are retrieved, validated using topology, resource, and outcome checks, and either reused or rejected in favor of a planner/reasoning fallback.}
    \label{fig:runtime-architecture}
    \vspace{-5mm}
\end{figure*}
\textbf{\textit{\underline{{(C) Runtime Objective:}}}} The runtime objective is to select a supervisory action during communication-limited execution. Given current state \(s_t\), memory bank \(\mathcal{M}\), and retrieved candidate set \(\mathcal{C}_t^k\), the robot should reuse a remembered action only if at least one candidate passes validation:
\begin{equation}
a_t =
\begin{cases}
a_i, & \exists m_i\in\mathcal{C}_t^k \text{ such that } \Phi(s_t,m_i)=1,\\
f(s_t), & \text{otherwise},
\end{cases}
\vspace{-1ex}
\end{equation}
where \(f(s_t)\) denotes the local planner, local reasoning fallback, or safe supervisory policy. Similarity-only reuse implicitly assumes that the highest-scoring memory is executable. \texttt{MemoGuard} instead validates candidate memories before reuse, rejecting memory traps whose execution contracts no longer hold.

\vspace{-4ex}
\section{\texttt{MemoGuard} Runtime}
\label{sec:method}

\texttt{MemoGuard} (Fig.~\ref{fig:runtime-architecture}) is a lightweight runtime layer between episodic memory retrieval
and action execution. Fig.~\ref{fig:runtime-architecture}(b) shows the episodic memory
format and Fig.~\ref{fig:runtime-architecture}(c) illustrates the runtime flow: given
the current robot state and memory bank, \texttt{MemoGuard} retrieves candidate
memories, validates their execution contracts, and either reuses a remembered
supervisory action or invokes a planner/reasoning fallback.\\
\noindent
\textbf{\textit{\underline{(A) Candidate Retrieval:}}} At each decision step, \texttt{MemoGuard} masks actions inconsistent with the current
mission phase and retrieves memories matching the current intent context, node, target,
and mission phase; if the robot is already at the target or a valid alternate viewpoint,
it selects \texttt{inspect\_target} directly.\\
Eligible memories are ranked using a retrieval score combining environmental and
agent-state similarity (Fig.~\ref{fig:runtime-architecture}(c), left):
\begin{equation}
S_{\mathrm{ret}}(s_t,m_i)
=
\lambda_{\mathrm{env}}S_{\mathrm{env}}(s_t,m_i)
+
\lambda_{\mathrm{agent}}S_{\mathrm{agent}}(s_t,m_i),
\vspace{-1ex}
\end{equation}
where \(S_{\mathrm{env}}\) and \(S_{\mathrm{agent}}\) are normalized to \([0,1]\)
and \(\lambda_{\mathrm{env}}+\lambda_{\mathrm{agent}}=1\) (we use
\(\lambda_{\mathrm{env}}=0.8\), \(\lambda_{\mathrm{agent}}=0.2\) to emphasize
local traversability). \(S_{\mathrm{env}}\) compares local visibility, obstacle state,
and edge-cost similarity; \(S_{\mathrm{agent}}\) compares battery range and
localization-confidence level.\\
\noindent
\textbf{\textit{\underline{{(B) Execution-Contract Validation:}}}} A retrieved memory is not reused solely because it is similar. As shown in
Fig.~\ref{fig:runtime-architecture}(c), \texttt{MemoGuard} applies three hard
feasibility gates as mentioned below (intent compatibility is enforced during eligibility filtering;
validation covers topology, resources, and outcome reliability).\\
\noindent
\textbf{Topology check.}
It rejects memories whose structural assumptions no longer hold, including cases where
the required edge is blocked or the remembered alternate viewpoint is unavailable.\\
\textbf{Resource check.}
It rejects memories that would drive the robot below the configured battery safety floor.
Let \(b_t\) be the current battery, \(C_{\mathrm{act}}(s_t,m_i)\) be the estimated
cost of executing the remembered action, and \(B_{\mathrm{safe}}\) be the safety floor.
The resource condition is
\begin{equation}
    b_t - C_{\mathrm{act}}(s_t,m_i) \geq B_{\mathrm{safe}}.
    \vspace{-1ex}
\end{equation}
If the remembered action target is unavailable or unreachable, the action cost is
treated as infeasible and the memory is rejected.
\\
\textbf{Outcome check.}
It rejects memories with unreliable prior outcomes. \texttt{MemoGuard} computes empirical
reliability as
\begin{equation}
    R(m_i)=\frac{n_i^s}{n_i^s+n_i^f},
    \vspace{-1ex}
\end{equation}
where \(n_i^s\) and \(n_i^f\) are accumulated success and failure counts for
equivalent state-action memories, and reuses only if \(R(m_i)\) exceeds the outcome
threshold \(\tau_{\mathrm{out}}\).\\
\noindent
\textbf{\textit{\underline{(C) Reuse or Fallback:}}}
A candidate memory is valid only if all checks pass:
\begin{equation}
\Phi(s_t,m_i)
=
\Phi_{\mathrm{topo}}(s_t,m_i)
\wedge
\Phi_{\mathrm{res}}(s_t,m_i)
\wedge
\Phi_{\mathrm{out}}(m_i).
\end{equation}
As shown in Fig.~\ref{fig:runtime-architecture}(c), if \(\Phi(s_t,m_i)=1\),
\texttt{MemoGuard} reuses \(a_i\); otherwise it checks the next candidate. If no
candidate passes, the runtime invokes fallback \(f(s_t)\), which selects one action
from \(\mathcal{A}\) (Eq. \ref{eq1}). The reasoning fallback is instantiated as local
\texttt{llama3.2:3b} inference constrained to a single supervisory action, following recent works on LLM-based high-level planning involving embodied agents~\cite{ahn2022saycan,10.1145/3676641.3716016, su2025data, tahmasbi2025building}.
\vspace{-2ex}
\section{Experiments and Evaluation}
\label{sec:evaluation}
\vspace{-0.5ex}
We evaluate whether \texttt{MemoGuard} reduces unsafe episodic memory reuse while
avoiding the cost of always-reasoning, using a controlled graph-world stress test where
retrieved memories may appear similar but become invalid due to topology, resource, or
alternate-viewpoint changes.


\textbf{\textit{\underline{(A) Setup:}}} We implement a graph-based corridor inspection simulator for a ground robot operating
over an initially available topological map. As shown in
Fig.~\ref{fig:runtime-architecture}(a), the simulator uses three map families:
\emph{Linear Corridor}, \emph{Alternate Path}, and \emph{Long Return Cost}. Edges
encode traversal cost and obstacle state; nodes encode local visibility.


Each scenario randomizes edge costs, obstacle states, visibility levels, target
location, alternate-viewpoint availability, initial battery, and localization
confidence, with a battery safety floor of \(B_{\mathrm{safe}}=25\) and a maximum
episode length of 50 steps. 


Movement actions use a graph planner computing minimum-cost paths via Dijkstra's
algorithm. For \texttt{follow\_planner}, the planner targets the inspection node before
inspection and the safe waypoint or base after; for \texttt{return\_to\_safe\_waypoint}
and \texttt{inspect\_alternate\_viewpoint}, it targets the corresponding recovery or
alternate-viewpoint node. The reasoning fallback returns one action from \(\mathcal{A}\) (Eq.~\ref{eq1})
and increments the fallback-call metric; graph planning only executes selected movement actions.


The memory bank is constructed offline from rollout data: successful transitions are
stored as episodic action memories accumulating success counts, and failed rollouts
contribute failure counts when they match an existing memory signature, yielding
11,558 memories. During evaluation, the retriever samples up to 2,000 memories using
a fixed seed and ranks them using the retrieval score from Section~\ref{sec:method}.

\textbf{\textit{\underline{{(B) Trap Generation, Baselines, and Metrics:}}}} To test memory traps, we generate trap scenarios from previously generated non-trap scenarios. Each trap applies one controlled modification while preserving the overall mission structure; only one trap type is applied per scenario.

\begin{itemize}
    \item \textbf{Blocked Edges}: one or more previously traversable edges are changed to blocked.
    \item \textbf{Reduced Battery}: the robot's initial battery is reduced by 15\%, subject to a minimum feasible battery level.
    \item \textbf{Removed Alternate Viewpoint}: the alternate viewpoint is removed and treated as a normal node with no alternate-inspection affordance.
\end{itemize}

Trap scenarios remain solvable after modification, so failures reflect invalid memory reuse rather than impossible missions.

We compare four policies. \textbf{Top-1 Reuse} directly executes the highest-scoring memory; \textbf{Threshold Reuse} adds a similarity threshold, invoking fallback if the score is too low. \textbf{Always Reasoning} invokes the local reasoning fallback at every decision step. \textbf{\texttt{MemoGuard}} retrieves a candidate memory and validates it before reuse.

We report six metrics. \emph{Mission Success} is the percentage of trials in which the robot inspects the target, preserves the battery safety floor, and terminates at a recovery node. \emph{Target Inspected} is the percentage of trials in which the target is inspected, regardless of final recovery status. \emph{Battery Depletion} is the percentage of trials in which the battery reaches zero. \emph{Battery Safety Violation} is the percentage of trials ending below the configured battery safety floor. \emph{Fallback Calls} is the average number of planner/reasoning fallback invocations per trial. \emph{Execution Cost} is the average battery-equivalent cost accumulated over movement, inspection, waiting, and fallback overhead.

\textbf{\textit{\underline{{(C) Memory-Trap Results:}}}} Table~\ref{tab:average_results} summarizes average performance across trap scenarios. Similarity-only reuse is cheap but unsafe: Top-1 Reuse has zero fallback calls, but only 32.6\% mission success and 67.4\% battery safety violations. Threshold Reuse performs similarly, with 67.1\% battery safety violations, showing that retrieval-score thresholding alone does not solve the memory-trap problem. Many traps are not low-confidence retrieval failures; they are high-similarity memories whose topology, resource, or alternate-viewpoint assumptions are invalid.

\begin{table}[t]
\centering
\caption{\scriptsize {Average performance across trap scenarios. Mission = mission success; Inspect = target inspected; Deplete = battery depletion; Safety Viol. = battery safety violation; Fallback = average fallback calls per trial; Cost = battery-equivalent execution cost. Arrows indicate preferred direction.}}
\label{tab:average_results}
\resizebox{\columnwidth}{!}{
\begin{tabular}{lrrrrrr}
\toprule
Policy & Mission $\textcolor{red}{\uparrow}$ & Inspect $\textcolor{red}{\uparrow}$ & Deplete $\textcolor{blue}{\downarrow}$ & Safety Viol. $\textcolor{blue}{\downarrow}$ & Fallback $\textcolor{blue}{\downarrow}$ & Cost $\textcolor{blue}{\downarrow}$ \\
\midrule
Top-1 Reuse & 32.6 & 80.4 & 18.9 & 67.4 & \textbf{0.00} & \textbf{36.73} \\
Threshold Reuse & 32.7 & 76.6 & 18.9 & 67.1 & 0.53 & 37.09 \\
Always Reasoning & 83.5 & \textbf{98.6} & 2.2 & 16.5 & 18.58 & 63.03 \\
\texttt{MemoGuard} & \textbf{84.2} & 98.2 & \textbf{2.0} & \textbf{15.8} & 14.60 & 60.21 \\
\bottomrule
\end{tabular}
}
\vspace{-4mm}
\end{table}

\begin{figure}[t]
    \centering
    \includegraphics[width=0.4\textwidth]{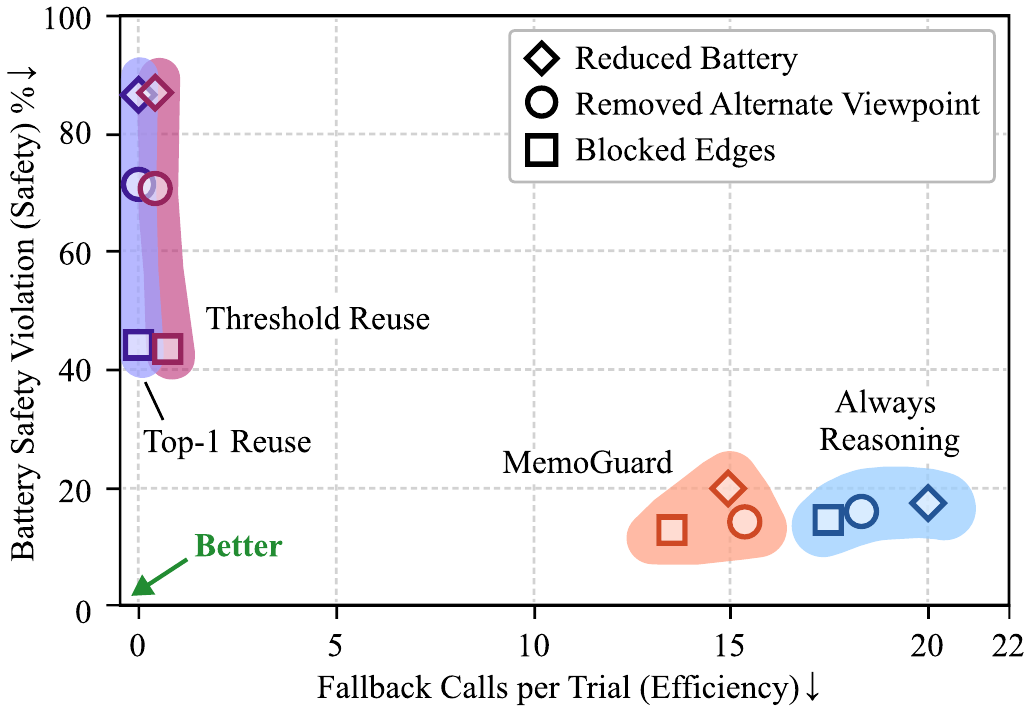}
    \vspace{-2ex}
    \caption{\scriptsize Safety--efficiency tradeoff by trap type. Each point is one policy averaged over one trap mechanism; lower-left is better. Text labels identify trap mechanisms and policy operating regions. \texttt{MemoGuard} approaches Always Reasoning safety with fewer fallback calls, with the largest gains for blocked-edge and alternate-viewpoint traps.}
    \label{fig:safety_efficiency}
    \vspace{-6mm}
\end{figure}


Compared with Top-1 Reuse, \texttt{MemoGuard} increases mission success from 32.6\% to 84.2\%. Overall, battery safety violations drop from 67.4\% to 15.8\%, corresponding to a 76.6\% relative reduction. Compared with Always Reasoning, \texttt{MemoGuard} reduces fallback calls from 18.58 to 14.60 per trial, a 21.4\% reduction, while maintaining comparable mission success: 84.2\% vs. 83.5\%. Thus, \texttt{MemoGuard} improves the safety-efficiency tradeoff: it is much safer than similarity-only reuse and less fallback-intensive than always reasoning.


Fig.~\ref{fig:safety_efficiency} disaggregates the safety--efficiency tradeoff by trap mechanism. In Blocked Edges scenarios, \texttt{MemoGuard} reduces safety violations from 44.3\% under Top-1 Reuse to 13.0\%, slightly below Always Reasoning at 15.0\%, while using fewer fallback calls: 13.51 vs. 17.47. In Removed Alternate Viewpoint scenarios, \texttt{MemoGuard} reduces violations from 71.2\% to 14.4\%, compared with 16.5\% for Always Reasoning, while reducing fallback calls from 18.28 to 15.37. The Reduced Battery case is harder: \texttt{MemoGuard} reduces violations from 86.8\% to 20.1\%, but remains above Always Reasoning at 17.9\%; however, it still uses fewer fallback calls, 14.94 vs. 19.99. These trends show that \texttt{MemoGuard} is strongest when memory traps arise from discrete topology or affordance changes, while resource-margin traps remain more sensitive to cost estimation.

On Jetson AGX Xavier in \texttt{MODE\_30W\_ALL}, local \texttt{llama3.2:3b} fallback reasoning costs 0.922~s and 9.288~J per call; therefore, \texttt{MemoGuard}'s 3.98-call reduction over Always Reasoning in Table~\ref{tab:average_results} corresponds to 3.67~s and 36.97~J of avoided fallback-reasoning overhead per trial. 


\vspace{-2mm}
\section{Conclusion and Future Work}
\vspace{-1ex}
This paper presents \texttt{MemoGuard}, a lightweight runtime for contract-validated episodic memory reuse in communication-limited robot navigation. \texttt{MemoGuard} addresses memory traps by validating a retrieved memory's topology, resource, and outcome assumptions before reuse. This validation shifts the safety-efficiency tradeoff away from unsafe similarity-only reuse and reduces reliance on costly always-reasoning. In corridor-inspection scenarios, \texttt{MemoGuard} reduced battery safety violations by 76.6\% over similarity-only top-1 reuse and fallback calls by 21.4\% over always reasoning. Future work will extend \texttt{MemoGuard} from graph-level contracts to cross-layer execution contracts that incorporate localization uncertainty, communication quality, and energy/resource trajectories for communication-constrained mobile robots~\cite{bhattacharjya2026access,wu2026joint, bera2}, extending toward multi-robot collaboration~\cite{riwa1, riwa2, zh, ravichandran2025heterogeneous}. 
\vspace{-2ex}

\bibliographystyle{IEEEtran}

\end{document}